\documentclass[letterpaper]{article} 
\usepackage[preprint]{aaai2027}  
\usepackage[hyphens]{url}  
\usepackage{graphicx} 
\usepackage{natbib}  
\usepackage{caption} 
\usepackage{algorithm}
\usepackage{algorithmic}

\usepackage{newfloat}
\usepackage{listings}
\DeclareCaptionStyle{ruled}{labelfont=normalfont,labelsep=colon,strut=off} 
\floatstyle{ruled}
\newfloat{listing}{tb}{lst}{}
\floatname{listing}{Listing}

\usepackage{amsmath}
\usepackage{amssymb}
\usepackage{amsfonts}
\usepackage{amsthm}
\usepackage{booktabs}
\usepackage{multirow}
\usepackage[table]{xcolor}
\usepackage{array}
\usepackage{textcomp}
\usepackage{makecell}
\usepackage{pifont}

\definecolor{grayrow}{gray}{0.94}
\definecolor{myred}{RGB}{200,0,0}
\definecolor{myblue}{RGB}{0,90,180}
\newcommand{\res}[2]{#1$_{\pm #2}$}
\newcommand{\imp}[1]{\textbf{+#1}}
\newcommand{\gain}[1]{\ensuremath{^{\textcolor{myblue}{+\,#1}}}}
\newcommand{\loss}[1]{\ensuremath{^{\textcolor{myred}{-\,#1}}}}

\usepackage[caption=false]{subfig}
\title{Dissecting Federated-Graph Aggregation under Domain Shift:\\ Importance-Aware Aggregation via Empirical Analysis}
\author{
    Zhanting Zhou\textsuperscript{\rm 1}\corresponding,
    Kahou Tam\textsuperscript{\rm 2},
    Zeyu Ma\textsuperscript{\rm 1},
    Ziqiang Zheng\textsuperscript{\rm 1}
}
\affiliations{
    \textsuperscript{\rm 1}University of Electronic Science and Technology of China\\
    \textsuperscript{\rm 2}State Key Laboratory of Internet of Things for Smart City, University of Macau\\
    ztzhou@std.uestc.edu.cn
}

\begin{document}

\maketitle

\begin{abstract}
Federated graph learning (FGL) trains a shared graph model across clients whose local graphs differ in node features, labels, and connectivity while keeping raw graph data decentralized. Although graph-domain shifts across clients can severely degrade the global model, existing FGL approaches for graph-domain shift mainly adapt local representations, propagation, or graph-derived collaboration, while the server typically applies the same aggregation rule to every parameter coordinate. In this work, we analyze how graph-domain shifts affect server-side aggregation. We find that, as clients optimize under distinct graph-domain conditions, they gradually concentrate their strongest updates on different parameter coordinates, making important update coordinates less shared across clients. Consequently, standard averaging can weaken these domain-specific signals even when client updates are not directly opposed. These findings reveal an aggregation-induced signal dilution effect: parameter coordinates strongly expressed by only a subset of domains are attenuated by near-zero contributions from the remaining domains. Motivated by these findings, we propose FedIA, a lightweight server-side aggregation method that calibrates parameter selection and client weighting. FedIA identifies important coordinates within each layer and adjusts client aggregation weights according to their contributions on the selected subspace, without modifying local graph training or the standard client-update payload. Experiments on Twitch Gamers and WikiNet using two graph-learning backbones and nine FL/FGL baselines show improvements of up to 11.58 percentage points, with particularly strong gains under domain skew.
\end{abstract}


\section{Introduction}
\label{sec:introduction}
\begin{figure*}[t]
    \centering
    \includegraphics[width=0.9\linewidth]{Figures/opening_v1.pdf}
    \caption{
    (a) Cross-silo clients learn one graph model from private graph domains and expose only model updates to the server.
    (b) Existing FGL research primarily intervenes in local graph learning or generic server optimization. We examine how graph-learning configurations shape client updates and how fragmented salient activity and weak cross-domain reinforcement affect server aggregation.
    (c) FedIA uses layer-wise Importance Masking to select parameters and Contribution-Aware Momentum to weight clients on the selected subspace before forming the global update.}
    \label{fig:opening}
\end{figure*}

Federated learning (FL)~\citep{fedavg} enables decentralized clients to train a shared global model without centralizing raw data, but heterogeneity among local data distributions remains a fundamental challenge~\citep{kairouz2021advances}. This challenge is especially pronounced in cross-silo graph settings, where each client commonly represents a persistent graph domain rather than a non-IID partition. These domains can differ in node features, labels, and connectivity patterns. Federated graph learning (FGL)~\citep{fgl1,fgl2,fgl3} addresses this setting by training a shared graph model across clients while keeping each client's graph data decentralized~\citep{fglsurvey25}. In practical cross-silo applications, such as hospital networks and organizational graphs~\citep{fedsage,fggp}, clients may contain complementary yet domain-specific graph patterns, making collaboration beneficial but vulnerable to feature, label, and structural heterogeneity~\citep{openfgl}.

Figure~\ref{fig:opening} summarizes how graph-domain shifts across clients reach server-side aggregation. These shifts are encoded into local updates, as client-specific feature, label, sampling, and structural conditions influence the gradients produced during local graph-model training. Client-specific graph conditions affect local updates differently across graph-learning paradigms. Message-passing graph neural networks (GNNs) use private adjacency during training through neighborhood aggregation, whereas propagation-deferred methods, \emph{e.g.}, PMLP~\citep{pmlp}, separate parameter learning from graph propagation by optimizing shared multilayer perceptron (MLP) weights locally and applying propagation only at inference. Across these paradigms, the server cannot inspect the private graph factors that produced an update. It only receives the update before forming the next global model. The received domain-specific signals vary across parameter coordinates, i.e., scalar dimensions of the model update. At aggregation, client relevance is therefore parameter-dependent: clients carrying strong signals for one coordinate may be weak or inactive for another. We use cross-domain aggregation compatibility to describe whether domain-specific updates activate shared coordinates and reinforce a common global step. Standard averaging assumes away this parameter dependence by assigning each client one global aggregation weight across all coordinates. Consequently, signals concentrated in a subset of relevant clients may be diluted by updates from less relevant domains, producing aggregation-induced signal dilution.

This aggregation-side mismatch complements the heterogeneity addressed by existing FGL methods. Existing FGL methods~\citep{fedsage,fedspray,fggp,fedspa,fedgcm,fedsst} act before or around aggregation through neighbor recovery, structural alignment, propagation modification, or graph-specific grouping and weighting. Beyond FGL, generic federated aggregation methods~\citep{gma,fedlaw,fedawa,fedheal} refine client weighting through directional agreement, learnable client weights, or whole-update alignment. Their principal signals remain auxiliary, representation-level, directional, or client-level. In this work, we ask whether the local model-update vectors uploaded after graph-model training already encode feature, label, sampling, and structural heterogeneity, and whether this parameter-wise evidence can correct server aggregation without additional client payloads.

We investigate this question through three complementary analyses of server aggregation. First, repeated local optimization under distinct graph-domain conditions makes the largest update components increasingly client-specific, revealing salient update fragmentation: strong signals concentrate on different parameter dimensions, weakening cross-client reinforcement and amplifying dilution under standard averaging. Second, because different domains often activate only partially overlapping parameter subspaces, cross-domain updates tend to align weakly with the global step: they provide limited positive reinforcement rather than coherent support, which slows useful signal accumulation during aggregation. Third, aggregation compatibility is induced after local graph learning through the interaction between model architecture and feature, label, sampling, and structural heterogeneity. It therefore emerges from fragmented parameter activity, weak cross-client reinforcement, and architecture-dependent alignment.

These observations identify a shared aggregation mismatch: client relevance varies across parameters, so each model component may be supported by a different effective contributor set. However, standard averaging collapses this parameter-dependent structure into a single client weight, causing each component to be mixed with weakly relevant updates and thereby producing aggregation-induced signal dilution. Motivated by this mechanism, we propose FedIA, a parameter-wise aggregation framework that derives aggregation importance from the graph-domain-conditioned activity encoded in client updates. Layer-wise Importance Masking (IM) selects high-activity parameters within each layer, defining the subspace where aggregation should be most selective. Contribution-Aware Momentum (CAM) then reweights clients by their contribution on this subspace and smooths the weights across rounds. FedIA therefore uses the graph-domain-conditioned signal already available at the server without requesting additional client payloads.


\begin{table*}[!t]
\centering
\caption{Comparison with existing methods. FedIA targets cross-silo FGL under graph-domain shifts in node classification, dissects the ordinary client-update stream, and preserves standard update exchange.}
\label{tab:method_comparison}
\resizebox{\linewidth}{!}{%
\begin{tabular}{llcccc}
\toprule
\textbf{Method Categories} & \textbf{Representative Methods} &
\textbf{\makecell{Cross-silo\\domain shift}} &
\textbf{\makecell{Graph node\\classification}} &
\textbf{\makecell{Update-level\\dissection}} &
\textbf{\makecell{Standard\\update exchange}} \\
\midrule
Generic heterogeneity FL
& FedProx, FedDyn
& \ding{55}
& \ding{55}
& \ding{55}
& \ding{51} \\

Prototype-based FGL
& FGSSL, FGGP
& \ding{51}
& \ding{51}
& \ding{55}
& \ding{55} \\

Homophily-aware FGL
& FedSPA
& \ding{51}
& \ding{51}
& \ding{55}
& \ding{51} \\

Domain-shift aggregation solution
& FedHEAL
& \ding{51}
& \ding{55}
& \ding{55}
& \ding{55} \\

\textbf{Ours}
& \textbf{FedIA}
& \ding{51}
& \ding{51}
& \ding{51}
& \ding{51} \\
\bottomrule
\end{tabular}%
}
\end{table*}

Experiments on Twitch Gamers~\citep{twitch} and WikiNet~\citep{wikinet}, using PMLP-GCN and GraphSAGE as representative decoupled-propagation and message-passing graph learners, show that FedIA consistently improves over nine FL and FGL baselines. FedIA improves average accuracy by up to 11.58 percentage points. The gains appear across domains and are especially pronounced under stronger domain skew, confirming the importance of correcting parameter-dependent dilution during aggregation. When combined with strong graph-specific alignment methods, FedIA yields smaller but still complementary improvements, suggesting a division of labor in the federated graph-learning pipeline: local graph methods improve domain-specific representation learning, while FedIA calibrates the client updates consolidated at the server. We summarize our contributions as follows:

\begin{itemize}
    \item \textbf{Parameter-level dissection of graph-domain aggregation.}
    We show that graph-domain heterogeneity manifests in client-uploaded updates as salient update fragmentation, weak cross-domain reinforcement, and architecture-dependent aggregation compatibility.

    \item \textbf{FedIA: update-driven parameter-wise aggregation.}
    We identify aggregation-induced signal dilution as the mismatch between parameter-dependent client relevance and parameter-invariant averaging, and propose FedIA to correct it using only uploaded updates.

    \item \textbf{Validation across graph domains and learners.}
    We validate FedIA on multi-domain graph benchmarks with decoupled-propagation and message-passing learners, showing consistent gains over heterogeneity-aware FL/FGL baselines, especially under domain skew, without altering local training or client payloads.
\end{itemize}

\section{Related Work}
\label{sec:related_works}

\begin{figure}[!t]
    \centering
    \includegraphics[width=\linewidth]{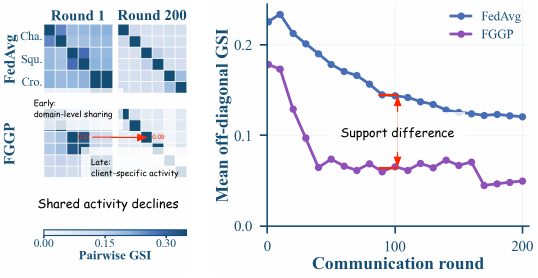}
    \caption{Training-time fragmentation of salient update coordinates on WikiNet. Pairwise top-$5\%$ Jaccard overlap retains visible domain blocks early in training, while these blocks weaken at later checkpoints. Mean off-diagonal overlap decreases under FedAvg and FGGP and remains above the random-set reference.}
    \label{fig:o1}
\end{figure}

\paragraph{Federated graph learning.}
FGL extends federated optimization to graph-structured data whose nodes, attributes, and edges are distributed across clients~\citep{openfgl,fgl1,fgl2,fgl3,fglsurvey25}.
Existing methods act at several points of the collaboration pipeline. FedSage+~\citep{fedsage} reconstructs missing cross-client neighbors, FedSpray~\citep{fedspray} aligns class-wise structure proxies, and FedGTA~\citep{li2024fedgta} uses topology-derived information for personalized aggregation.
FGSSL~\citep{fgssl} calibrates node semantics and graph relations during local learning.
These methods establish graph structure as an essential federated signal, while their primary objects remain neighborhoods, representations, structural proxies, and collaboration topology.

\paragraph{Graph-domain and structural shifts.}
The closest FGL line studies clients whose local graphs arise from different domains or structural regimes.
FGGP~\citep{fggp} learns generalizable prototypes across participating graph domains.
FedSPA~\citep{fedspa} addresses homophily conflict and bias through propagation decoupling and homophily-driven aggregation.
FedGCM~\citep{fedgcm} organizes homophily regimes and applies gradient surgery to reduce inter-group interference, while FedSST~\citep{fedsst} uses a structure-derived signal for fairness-aware aggregation and adaptive local training.
These methods dissect graph-domain shift through representations, propagation, client groups, and explicit structural signals.
FedIA extends this line to the server-side by studying the ordinary client updates produced after local graph learning.

\begin{figure}[!t]
    \centering
    \includegraphics[width=\linewidth]
    {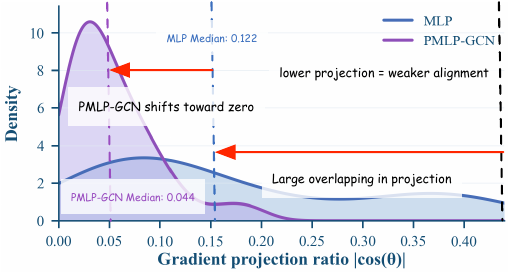}
    \caption{
    Configuration sensitivity of cross-domain update alignment.
    Under identical graph-domain partitions, PMLP-GCN shifts pairwise alignment magnitude $|\cos\theta|$ toward zero relative to MLP.
    Since PMLP optimizes its shared weights without propagation, this comparison characterizes the server-visible update field in FGL.
    }
    \label{fig:o2}
\end{figure}

\paragraph{Update-aware aggregation.}
Generic FL provides direct algorithmic neighbors at the server.
Gradient Masked Averaging (GMA)~\citep{gma} selects coordinates according to directional agreement.
FedLAW~\citep{fedlaw} learns aggregation weights from client coherence with a server-side proxy set, whereas FedAWA~\citep{fedawa} adapts client weights from client vectors and their relation to the global update.
FedHEAL~\citep{fedheal} tracks parameter-update consistency, selects important coordinates, and adjusts aggregation under domain skew with a fairness objective and client-specific coordinate histories.
FedIA begins from a complementary aggregation dissection: salient activity becomes less shared, cross-domain updates weakly reinforce the aggregate, and the active contributor set varies across model components.
It therefore selects a layer-wise current-round parameter subspace before estimating and smoothing client contributions on that same subspace.
Table~\ref{tab:method_comparison} summarizes the graph-domain scope, node-classification setting, update-level dissection, and update-exchange interface of the representative methods.

\section{Importance-aware Aggregation via Empirical Analysis}
\label{sec:method}


\begin{figure}[!t]
    \centering
    \includegraphics[width=\linewidth]{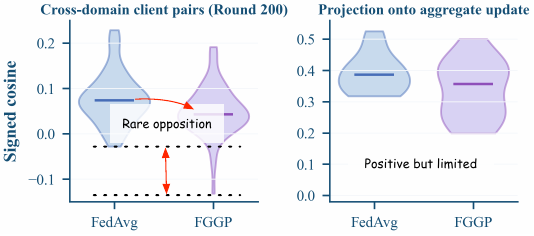}
    \caption{Directional contribution to aggregation. Pairwise signed cosine identifies opposing client updates, while $\cos(u_k,\bar{u})$ measures each client's projection onto the update produced by ordinary averaging. Together, the two views quantify weak cross-domain reinforcement of the global step.}
    \label{fig:o3}
\end{figure}

\subsection{Client Updates as Server-Observable Signals}\label{sec:update_view}

At round $t$, client $k\in\mathcal S_t$ returns an update
$u_k^t\in\mathbb R^D$.
For gradient sharing, $u_k^t$ is the uploaded gradient; for model sharing, we express the negative model delta in the same descent convention as
$u_k^t=-\Delta W_k^t/\eta$.
After multiple local steps, $u_k^t$ represents the realized update aggregated by the server.
Stacking $|u_k^t|$ across clients yields a client--parameter activity pattern: each row records where one graph domain changes the model, while each column records how broadly a model component is activated.

For a training-time message-passing GNN, local graph conditions enter the update through
\begin{equation}
g_k^t =
\sum_{v\in\mathcal V_k}
\frac{\partial \mathcal L_k}{\partial h_v}
\frac{\partial h_v(A_k,X_k;W)}{\partial W}.
\end{equation}
The two factors transmit local supervision through feature activation and neighborhood aggregation. The uploaded update therefore provides the server with a common observation of feature, label, and structural effects
after local graph training.

\subsection{Findings on Aggregation Compatibility}
\label{sec:findings}

\begin{figure}[!t]
    \centering
    \includegraphics[width=\linewidth]{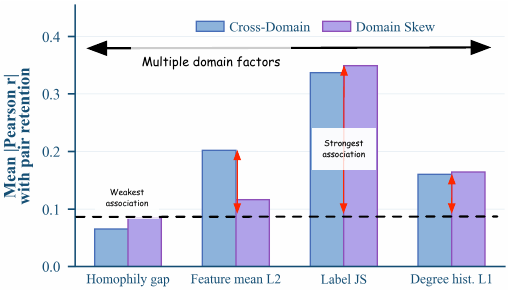}
    \caption{Descriptive associations between graph-domain discrepancies and salient-coordinate retention. Feature, label, and degree discrepancies show stronger associations than homophily gap across the tested backbones, supporting a multi-factor view of aggregation compatibility.}
    \label{fig:homo_weak}
\end{figure}

\noindent\textbf{Finding 1: salient update sharing erodes during training.}
A shared model depends on graph domains reinforcing at least part of the same parameter surface, so we track whether their strongest update components remain shared as optimization proceeds.
Let $S_k^t$ contain the top-$p\%$ coordinates of $|u_k^t|$ and define their pairwise Jaccard overlap as
\begin{equation}
 J_t(i,j)=\frac{|S_i^t\cap S_j^t|}{|S_i^t\cup S_j^t|}.
 \label{eq:salient_jaccard}
\end{equation}
This quantity is referred to as Gradient Support Intersection (GSI) in our figures.
With $p=5\%$, the random-set reference is $p/(2-p)=0.026$.
Figure~\ref{fig:o1} shows that within-domain blocks weaken and mean off-diagonal overlap declines under both FedAvg and FGGP.
The erosion accumulates throughout training, indicating that persistent domain-specific optimization increasingly concentrates high-magnitude activity in different model components.
This \textbf{salient update fragmentation} narrows the parameter surface through which graph domains jointly shape the global model.

\noindent\textbf{Finding 2: cross-domain updates weakly reinforce the global step.}
Shared coordinates can still cancel or reinforce one another, so overlap alone does not determine the quality of the aggregate.
Figure~\ref{fig:o2} shows that the recorded cross-domain alignment magnitudes concentrate near zero, and Figure~\ref{fig:o3} resolves their signs and projection onto the ordinary aggregate.
Only $5.0\%$ of FedAvg and $8.3\%$ of FGGP cross-domain pairs are negative at Round 200, while mean client-to-aggregate cosine remains positive but limited ($0.392$ and $0.339$).
The dominant pattern is therefore weak reinforcement: domains rarely push the global step in directly opposite directions, yet their updates share too little direction and concentrated activity to reinforce it strongly.
This finding motivates an activity-aware correction that preserves active components and calibrates contributor strength.

\noindent\textbf{Finding 3: aggregation compatibility depends on graph-learning configuration and domain factors.}
We next explore whether the observed compatibility is fixed by the domain partition or shaped by the complete graph-learning pipeline. Under identical partitions, the MLP/PMLP-GCN contrast in Figure~\ref{fig:o2} produces different overlap and alignment patterns.
Because PMLP-GCN defers propagation to inference, we use this result as evidence of configuration sensitivity in the server-visible update field.
We further compare homophily, feature means, label distributions, and degree histograms with the retention of locally salient components after aggregation.
Across five backbones, label divergence, feature distance, and degree discrepancy show stronger mean absolute associations than homophily gap as shown in Figure~\ref{fig:homo_weak}.
Aggregation compatibility is therefore formed downstream of the graph-learning configuration and several interacting domain factors.
The ordinary update stream provides a compact post-computation summary of this combined effect without requesting one preselected graph statistic from clients.


\begin{figure*}[!t]
    \centering
    \includegraphics[width=\linewidth]{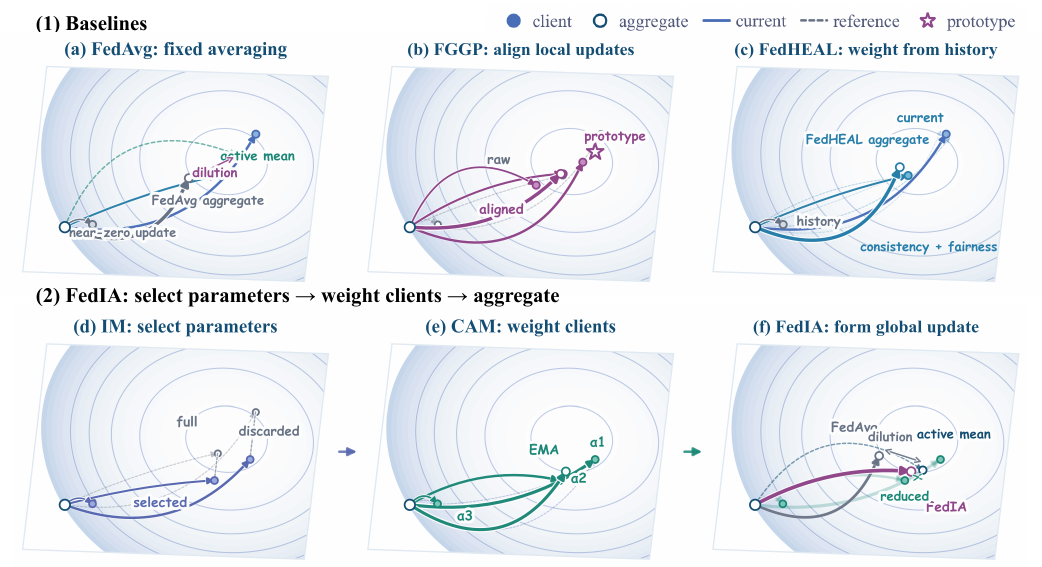}
    \caption{Layer-wise view of FedIA. IM selects high-importance coordinates within each predefined parameter layer. CAM then measures client activity on the selected coordinates and smooths the corresponding weights.}
\end{figure*}

\noindent\textbf{Aggregation consequence: parameter-dependent participation causes signal dilution.}
The three findings describe one aggregation mechanism.
Each graph domain produces a distinct row of client--parameter activity, so the effective contributor set changes across model components, while FedAvg applies one client-weight vector to every component.
Let $A_d^t$ be the clients for which component $d$ is salient and let $q_k$ be the base aggregation weight.
Their active share and active-client mean are
\begin{equation}
 r_d^t=\sum_{k\in A_d^t}q_k,\qquad
 \mu_{d,\mathrm{act}}^t=\frac{1}{r_d^t}\sum_{k\in A_d^t}q_k u_{k,d}^t.
 \label{eq:active_share}
\end{equation}
When the remaining clients are near zero on component $d$, ordinary averaging gives
\begin{equation}
 \bar{u}_d^t=\sum_k q_k u_{k,d}^t\approx r_d^t\mu_{d,\mathrm{act}}^t.
 \label{eq:signal_dilution}
\end{equation}
The aggregate is therefore scaled by the mass of clients that express the component, even when their active updates agree in sign.
We call this effect \textbf{aggregation-induced signal dilution}.
It motivates two coordinated server decisions: identify the parameter components carrying current activity, then stabilize the clients contributing on that selected subspace.

\subsection{FedIA: Importance-Aware Aggregation}
\label{sec:fedia}

\begin{algorithm}[!t]
\caption{Federated Importance-Aware Aggregation}
\label{alg:fedia}
\begin{algorithmic}[1]
\STATE \textbf{Input:} Initial model $W^0$, learning rate $\eta$, clients $\mathcal{C}$, participating subset $\mathcal{S}_t$, parameter groups $\mathcal{G}=\{G_m\}_{m=1}^M$, retention ratio $\rho$, and momentum factor $\beta$.
\STATE Initialize client weights $\alpha_k^0 \gets 1/|\mathcal{C}|$ for $k\in\mathcal{C}$.
\FOR{each round $t=0,1,\ldots,T-1$}
    \FOR{each parameter group $G_m\in\mathcal{G}$}
        \STATE $I_{G_m}^{t+1}\gets |\mathcal{S}_t|^{-1}\sum_{k\in\mathcal{S}_t}|u_{k,G_m}^{t+1}|$.
        \STATE $M_{G_m}^{t+1}\gets \mathrm{TopRatioMask}(I_{G_m}^{t+1},\rho)$.
    \ENDFOR
    \STATE $M^{t+1}\gets \mathrm{Concat}_{m=1}^{M} M_{G_m}^{t+1}$.
    \STATE $s_k^{t+1}\gets\|u_k^{t+1}\odot M^{t+1}\|_2$ for $k\in\mathcal{S}_t$.
    \STATE $p_k^{t+1}\gets \exp(s_k^{t+1})/\sum_{j\in\mathcal{S}_t}\exp(s_j^{t+1})$.
    \STATE $\tilde{\alpha}_k^{t+1}\gets\beta\alpha_k^t+(1-\beta)p_k^{t+1}$.
    \STATE $\alpha_k^{t+1}\gets\tilde{\alpha}_k^{t+1}/\sum_{j\in\mathcal{S}_t}\tilde{\alpha}_j^{t+1}$.
    \STATE $u^{t+1}\gets\sum_{k\in\mathcal{S}_t}\alpha_k^{t+1}(u_k^{t+1}\odot M^{t+1})$.
    \STATE $W^{t+1}\gets W^t-\eta u^{t+1}$.
\ENDFOR
\STATE \textbf{return} $W^T$.
\end{algorithmic}
\end{algorithm}

FedIA combines a parameter mask and a client weight in the effective aggregation coefficient, $\omega_{k,d}^t=\alpha_k^t M_d^t$.
Importance Masking (IM) selects high-activity coordinates within predefined parameter layers, while Contribution-Aware Momentum (CAM) measures and smooths client activity on the selected subspace.
The method consumes only current client updates and leaves the client objective and graph-learning backbone unchanged.

```latex
\begin{table*}[!t]
\centering
\caption{Accuracy comparison on the \textbf{Twitch Gamers} severe domain-skew setting (ratio $1{:}10{:}1{:}1{:}1{:}1$, 30 clients). Results are reported as mean$\pm$std accuracy (\%). Superscripts denote gains and losses relative to the corresponding base method, including the average change in the AVG column. The FedIA (repl.) row replaces FedSPA's native aggregation rule.}
\label{tab:twitch_domain_skew}
\resizebox{\linewidth}{!}{\begin{tabular}{lcccccc|c}
\toprule
\multirow{2}{*}{\textbf{Method}} & \multicolumn{6}{c}{\textbf{Domains}} & \multicolumn{1}{c}{\textbf{Overall}}\\
\cmidrule(lr){2-7} \cmidrule(lr){8-8}
 & ES & DE & FR & RU & EN & PT & AVG\\
\midrule
\multicolumn{8}{c}{\textbf{PMLP-GCN}}\\
\midrule
FedAvg & \res{66.98}{0.06} & \res{38.76}{0.01} & \res{59.38}{0.05} & \res{65.06}{0.02} & \res{48.87}{0.03} & \res{59.97}{0.10} & \res{56.50}{0.04}\\
\quad\textbf{+ FedIA} & \res{71.29}{0.00}\textsuperscript{+4.31} & \res{38.95}{0.00}\textsuperscript{+0.19} & \res{62.75}{0.00}\textsuperscript{+3.37} & \res{76.40}{0.00}\textsuperscript{+11.34} & \res{45.53}{0.00}\textsuperscript{$-$3.34} & \res{64.02}{0.00}\textsuperscript{+4.05} & \res{59.82}{0.00}\textsuperscript{+3.32}\\
FedProx & \res{66.45}{0.00} & \res{38.47}{0.00} & \res{59.63}{0.00} & \res{65.55}{0.03} & \res{48.02}{0.01} & \res{59.48}{0.00} & \res{56.27}{0.01}\\
\quad\textbf{+ FedIA} & \res{69.98}{0.18}\textsuperscript{+3.53} & \res{50.57}{0.15}\textsuperscript{+12.10} & \res{62.56}{0.17}\textsuperscript{+2.93} & \res{73.29}{0.16}\textsuperscript{+7.74} & \res{55.37}{0.47}\textsuperscript{+7.35} & \res{68.85}{0.37}\textsuperscript{+9.37} & \res{63.44}{0.11}\textsuperscript{+7.17}\\
MOON & \res{66.95}{0.02} & \res{38.76}{0.01} & \res{59.56}{0.00} & \res{65.52}{0.02} & \res{48.58}{0.03} & \res{59.74}{0.00} & \res{56.52}{0.01}\\
\quad\textbf{+ FedIA} & \res{69.95}{0.06}\textsuperscript{+3.00} & \res{43.33}{0.03}\textsuperscript{+4.57} & \res{62.42}{0.07}\textsuperscript{+2.86} & \res{71.56}{0.04}\textsuperscript{+6.04} & \res{51.49}{0.04}\textsuperscript{+2.91} & \res{63.58}{0.05}\textsuperscript{+3.84} & \res{60.39}{0.03}\textsuperscript{+3.87}\\
FedDyn & \res{50.00}{21.29} & \res{50.00}{11.05} & \res{50.00}{14.02} & \res{50.03}{26.37} & \res{50.00}{4.47} & \res{50.00}{12.75} & \res{50.01}{9.81}\\
\quad\textbf{+ FedIA} & \res{71.29}{0.00}\textsuperscript{+21.29} & \res{38.95}{0.00}\textsuperscript{$-$11.05} & \res{62.75}{0.00}\textsuperscript{+12.75} & \res{76.40}{0.00}\textsuperscript{+26.37} & \res{45.50}{0.01}\textsuperscript{$-$4.50} & \res{64.02}{0.00}\textsuperscript{+14.02} & \res{59.82}{0.00}\textsuperscript{+9.81}\\
FedProto & \res{67.11}{0.09} & \res{38.78}{0.01} & \res{59.49}{0.06} & \res{65.15}{0.03} & \res{48.92}{0.02} & \res{59.87}{0.00} & \res{56.55}{0.02}\\
\quad\textbf{+ FedIA} & \res{62.42}{0.06}\textsuperscript{$-$4.69} & \res{54.40}{0.05}\textsuperscript{+15.62} & \res{60.48}{0.02}\textsuperscript{+0.99} & \res{63.50}{0.20}\textsuperscript{$-$1.65} & \res{56.99}{0.06}\textsuperscript{+8.07} & \res{61.30}{0.04}\textsuperscript{+1.43} & \res{59.85}{0.05}\textsuperscript{+3.30}\\
FGSSL & \res{67.19}{0.12} & \res{38.97}{0.02} & \res{59.58}{0.04} & \res{65.36}{0.04} & \res{50.01}{0.04} & \res{59.35}{0.00} & \res{56.74}{0.02}\\
\quad\textbf{+ FedIA} & \res{68.40}{0.03}\textsuperscript{+1.21} & \res{51.72}{0.03}\textsuperscript{+12.75} & \res{60.46}{0.02}\textsuperscript{+0.88} & \res{69.10}{0.02}\textsuperscript{+3.74} & \res{54.03}{0.03}\textsuperscript{+4.02} & \res{63.39}{0.04}\textsuperscript{+4.04} & \res{61.18}{0.01}\textsuperscript{+4.44}\\
FGGP & \res{63.51}{10.49} & \res{65.73}{1.09} & \res{54.71}{6.60} & \res{61.21}{13.15} & \res{55.67}{2.30} & \res{63.36}{8.55} & \res{60.70}{6.03}\\
\quad\textbf{+ FedIA} & \res{72.05}{1.28}\textsuperscript{+8.54} & \res{51.84}{1.74}\textsuperscript{$-$13.89} & \res{63.17}{1.85}\textsuperscript{+8.46} & \res{73.18}{2.17}\textsuperscript{+11.97} & \res{58.45}{2.11}\textsuperscript{+2.78} & \res{68.01}{0.57}\textsuperscript{+4.65} & \res{64.45}{0.38}\textsuperscript{+3.75}\\
FedSPA & \res{61.37}{0.02} & \res{51.81}{0.02} & \res{52.77}{0.02} & \res{63.30}{0.03} & \res{60.60}{0.04} & \res{58.80}{0.02} & \res{58.13}{0.04}\\
\quad\textbf{FedIA (repl.)} & \res{61.88}{0.03}\textsuperscript{+0.51} & \res{51.86}{0.02}\textsuperscript{+0.05} & \res{58.69}{0.01}\textsuperscript{+5.92} & \res{62.60}{0.04}\textsuperscript{$-$0.70} & \res{56.24}{0.02}\textsuperscript{$-$4.36} & \res{59.50}{0.03}\textsuperscript{+0.70} & \res{58.45}{0.04}\textsuperscript{+0.32}\\
\quad\textbf{+ FedIA} & \res{65.29}{0.23}\textsuperscript{+3.92} & \res{50.95}{0.02}\textsuperscript{$-$0.86} & \res{61.02}{0.10}\textsuperscript{+8.25} & \res{70.40}{0.24}\textsuperscript{+7.10} & \res{51.53}{0.09}\textsuperscript{$-$9.07} & \res{62.75}{0.04}\textsuperscript{+3.95} & \res{59.80}{0.14}\textsuperscript{+1.67}\\
\midrule
\multicolumn{8}{c}{\textbf{GraphSAGE}}\\
\midrule
FedAvg & \res{31.29}{0.25} & \res{61.05}{0.06} & \res{37.31}{0.48} & \res{25.39}{0.35} & \res{54.05}{0.16} & \res{38.08}{0.25} & \res{41.19}{0.20}\\
\quad\textbf{+ FedIA} & \res{39.38}{0.51}\textsuperscript{+8.09} & \res{43.18}{0.30}\textsuperscript{$-$17.87} & \res{37.98}{0.18}\textsuperscript{+0.67} & \res{46.07}{0.64}\textsuperscript{+20.68} & \res{51.67}{0.07}\textsuperscript{$-$2.38} & \res{47.52}{0.83}\textsuperscript{+9.44} & \res{44.30}{0.27}\textsuperscript{+3.11}\\
FedSage & \res{70.70}{0.00} & \res{38.95}{0.00} & \res{63.23}{0.03} & \res{75.66}{0.00} & \res{44.79}{0.02} & \res{62.75}{0.00} & \res{59.35}{0.01}\\
\quad\textbf{+ FedIA} & \res{71.30}{0.02}\textsuperscript{+0.60} & \res{38.95}{0.01}\textsuperscript{0.00} & \res{64.02}{0.00}\textsuperscript{+0.79} & \res{76.39}{0.03}\textsuperscript{+0.73} & \res{45.53}{0.01}\textsuperscript{+0.74} & \res{62.75}{0.00}\textsuperscript{0.00} & \res{59.82}{0.01}\textsuperscript{+0.47}\\
\bottomrule
\end{tabular}}
\end{table*}
```

\noindent\textbf{Importance Masking: selecting layer-wise parameters.}
Let $\mathcal{G}=\{G_m\}_{m=1}^M$ partition the model coordinates by trainable layer or predefined module.
For each group, the server computes
\begin{equation}
 I_{G_m}^t=\frac{1}{|\mathcal{S}_t|}\sum_{k\in\mathcal{S}_t}|u_{k,G_m}^t|,
 \qquad
 M_{G_m}^t=\operatorname{Top}_{\rho |G_m|}(I_{G_m}^t),
 \label{eq:group_importance}
\end{equation}
and concatenates the group masks as $M^t=\operatorname{Concat}_{m=1}^{M}M_{G_m}^t$ before projecting each update, $\hat{u}_k^t=u_k^t\odot M^t$.
Absolute magnitude summarizes parameter activity before signs interact, while the layer-wise budget prevents a small number of large layers from monopolizing selection.
IM therefore constructs a shared aggregation surface from the strongest current-round activity in each layer.
Clients continue to transmit the full update of the base protocol; selection occurs only at the server.

\begin{figure*}[t]
\centering
\begin{minipage}[t]{0.31\textwidth}
\centering
\includegraphics[width=\linewidth]{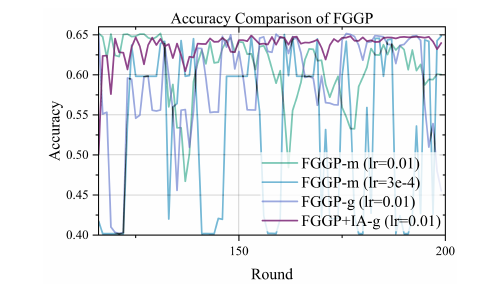}
\captionof{figure}{FGGP convergence. FedIA reduces late-round volatility under both update-communication modes.}
\label{fig:fggp_convergence}
\end{minipage}\hfill
\begin{minipage}[t]{0.31\textwidth}
\centering
\includegraphics[width=\linewidth]{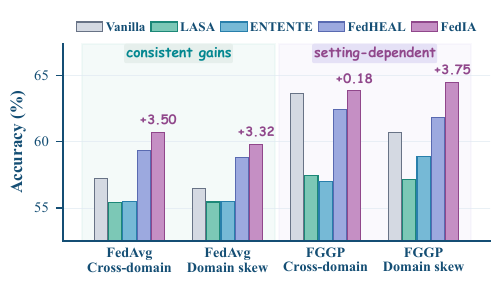}
\captionof{figure}{Aggregation comparison. FedIA consistently helps FedAvg, whereas FGGP exposes the setting-dependent.}
\label{fig:aggregation_comparison}
\end{minipage}\hfill
\begin{minipage}[t]{0.31\textwidth}
\centering
\includegraphics[width=\linewidth]{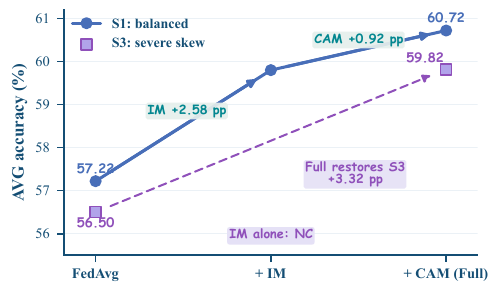}
\captionof{figure}{Ablations study. IM supplies most of the balanced gain; CAM is necessary for convergence under skew.}
\label{fig:component_ablation}
\end{minipage}
\end{figure*}

\begin{figure}[t]
\centering
\includegraphics[width=\linewidth]{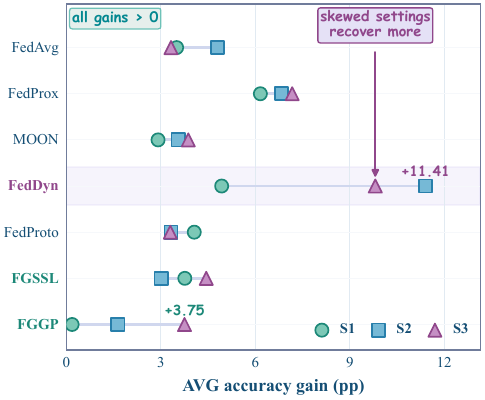}
\caption{FedIA's performance on Twitch Gamers. Markers report the AVG accuracy gain over each original baseline under S1, S2, and S3.}
\label{fig:fedia_gain_profile}
\end{figure}

\noindent\textbf{Contribution-Aware Momentum: weighting clients on the selected subspace.}
CAM scores each client on the selected subspace,
\begin{equation}
 s_k^t=\|\hat{u}_k^t\|_2,\qquad
 p_k^t=\frac{\exp(s_k^t)}{\sum_{j\in\mathcal{S}_t}\exp(s_j^t)},
 \label{eq:cam_score}
\end{equation}
and smooths these current-round weights with an exponential moving average,
\begin{equation}
 \tilde{\alpha}_k^t=\beta\alpha_k^{t-1}+(1-\beta)p_k^t,
 \qquad
 \alpha_k^t=\frac{\tilde{\alpha}_k^t}{\sum_{j\in\mathcal{S}_t}\tilde{\alpha}_j^t}.
 \label{eq:cam_weight}
\end{equation}
The global update is
\begin{equation}
 u^t=\sum_{k\in\mathcal{S}_t}\alpha_k^t\hat{u}_k^t,
 \qquad W^{t+1}=W^t-\eta u^t.
 \label{eq:fedia_update}
\end{equation}
Computing the score after masking ties each client weight to the parameter surface that is actually aggregated.
IM identifies the active parameter surface, while CAM suppresses round-to-round volatility in the client mixture operating on that surface.
The masked norm is an activity signal for aggregation, and the momentum state preserves temporal continuity across rounds.
Meanwhile, FedIA introduces no auxiliary client payload.
Its persistent server state consists of a $D$-dimensional profile or mask and $N$ scalar momentum weights, giving $\mathcal{O}(D+N)$ state; complete implementation details are provided in the supplementary material.

\section{Experiments}\label{sec:experiments}

\noindent\textbf{Experimental Settings.}
We evaluate node classification on Twitch Gamers~\citep{twitch} and WikiNet~\citep{twitch}, covering language-specific social networks and Wikipedia hyperlink networks with distinct feature, label, and connectivity regimes. The main study uses propagation-deferred PMLP-GCN~\citep{pmlp} with nine FL/FGL pipelines~\citep{fedavg,fedprox,moon,feddyn,fedproc,fedproto,fgssl,fggp,fedspa}; GraphSAGE~\citep{graphsage} with FedSage~\citep{fedsage} tests transfer to training-time neighborhood aggregation. Twitch S1/S2/S3 expand the DE allocation from $1{:}1$ to $5{:}1$ and $10{:}1$ relative to each other domain, while WikiNet uses a balanced $1{:}1{:}1$ split. We report mean$\pm$std accuracy. Appendix~\ref{app:experiments} specifies datasets, partitions, optimization, baselines, and the $\rho/\beta$ search; Appendix~\ref{app:full_results} gives full S1/S2 per-domain tables; Appendix~\ref{app:additional_results} gives the IM/CAM ablation, sensitivity, selected hyperparameters, and gradient inversion stress test.

\begin{table}[!t]
\centering
\captionof{table}{Accuracy comparison on \textbf{WikiNet} under the balanced cross-domain setting (ratio $1{:}1{:}1$, 6 clients). Results are reported as mean accuracy (\%) (without std report due to space limittation). Superscripts denote gains and losses relative to the corresponding base method, including the average change in the AVG column. The FedIA (repl.) row replaces FedSPA's native aggregation rule.}
\label{tab:wikinet_cross_domain}
\resizebox{\linewidth}{!}{\begin{tabular}{lccc|c}
\toprule
\multirow{2}{*}{\textbf{Method}} & \multicolumn{3}{c|}{\textbf{Domains}} & \multicolumn{1}{c}{\textbf{Overall}}\\
\cmidrule(lr){2-4} \cmidrule(lr){5-5}
 & Cha. & Cro. & Squ. & AVG\\
\midrule
FedAvg & 12.77 & 7.15 & 12.45 & 10.79\\
\quad\textbf{+ FedIA} & 25.02\textsuperscript{+12.25} & 20.55\textsuperscript{+13.40} & 21.54\textsuperscript{+9.09} & \textbf{22.37}\textsuperscript{+11.58}\\
FedProx & 12.06 & 5.72 & 10.14 & 9.31\\
\quad\textbf{+ FedIA} & 14.31\textsuperscript{+2.25} & 4.21\textsuperscript{$-$1.51} & 13.44\textsuperscript{+3.30} & \textbf{10.65}\textsuperscript{+1.34}\\
MOON & 12.72 & 6.83 & 12.56 & 10.70\\
\quad\textbf{+ FedIA} & 16.54\textsuperscript{+3.82} & 10.13\textsuperscript{+3.30} & 14.28\textsuperscript{+1.72} & \textbf{13.65}\textsuperscript{+2.95}\\
FedDyn & 15.45 & 8.50 & 13.46 & 12.47\\
\quad\textbf{+ FedIA} & 20.41\textsuperscript{+4.96} & 14.22\textsuperscript{+5.72} & 19.33\textsuperscript{+5.87} & \textbf{17.98}\textsuperscript{+5.51}\\
FedProc & 10.55 & 12.61 & 6.75 & 12.29\\
\quad\textbf{+ FedIA} & 20.04\textsuperscript{+9.49} & 10.34\textsuperscript{$-$2.27} & 19.96\textsuperscript{+13.21} & \textbf{16.78}\textsuperscript{+4.49}\\
FedProto & 12.65 & 7.16 & 12.48 & 10.76\\
\quad\textbf{+ FedIA} & 21.46\textsuperscript{+8.81} & 15.07\textsuperscript{+7.91} & 21.25\textsuperscript{+8.77} & \textbf{19.26}\textsuperscript{+8.50}\\
FGSSL & 14.45 & 7.97 & 13.64 & 12.02\\
\quad\textbf{+ FedIA} & 14.50\textsuperscript{+0.05} & 7.92\textsuperscript{$-$0.05} & 13.74\textsuperscript{+0.10} & \textbf{12.12}\textsuperscript{+0.10}\\
FGGP & 29.25 & 42.79 & 14.78 & 28.94\\
\quad\textbf{+ FedIA} & 42.99\textsuperscript{+13.74} & 14.86\textsuperscript{$-$27.93} & 29.21\textsuperscript{+14.43} & \textbf{29.02}\textsuperscript{+0.08}\\
FedSPA & 37.46 & 23.34 & 38.51 & 33.10\\
\quad\textbf{FedIA (repl.)} & 37.34\textsuperscript{$-$0.12} & 23.18\textsuperscript{$-$0.16} & 38.51\textsuperscript{0.00} & 33.01\textsuperscript{$-$0.09}\\
\quad\textbf{+ FedIA} & 37.44\textsuperscript{$-$0.02} & 23.49\textsuperscript{+0.15} & 38.57\textsuperscript{+0.06} & \textbf{33.17}\textsuperscript{+0.07}\\
\bottomrule
\end{tabular}}
\end{table}

\noindent\textbf{Paired effectiveness.}
Under severe Twitch skew, FedIA improves all eight PMLP-GCN averages by $0.32$--$9.81$ points and both GraphSAGE pairs by $0.47$--$3.11$ (Table~\ref{tab:twitch_domain_skew}). All nine WikiNet integrations also improve, up to $+11.58$ (Table~\ref{tab:wikinet_cross_domain}). The gains span the federation: 40/48 severe-skew PMLP-GCN cells (83.3\%) and 22/27 WikiNet cells (81.5\%) improve, and every PMLP-GCN pair gains on at least four Twitch domains. FedAvg's 3.34-point EN loss and FedProto's ES/RU losses reveal the remaining domain redistribution.

\noindent\textbf{Skew recovery.}
S3 replicates DE clients, yet 35/40 non-DE cells improve (87.5\%), including $+11.34$ on RU for FedAvg and $+9.37$ on PT for FedProx; DE improves for five of eight pairs. This matches signal dilution: DE replication reduces the active share $r_d^t$ of minority-domain coordinates, while IM preserves federation-wide activity and CAM smooths client weights. Figure~\ref{fig:fedia_gain_profile} shows positive S1--S3 gains and highlights FedDyn: its S1 result is 58.27$\pm$0.14, but S2/S3 fall to 50.01$\pm$9.79/9.81; FedIA restores 61.42$\pm$0.31/59.82$\pm$0.00 (Appendix~\ref{app:full_results}). Near-identical S3 FedDyn+FedIA and FedAvg+FedIA rows indicate that the selected subspace dominates the unstable dynamic correction in this setting. GraphSAGE adds a transfer boundary: FedIA gains 3.11 points on FedAvg but 0.47 on FedSage, whose neighbor generation already removes part of the mismatch.

\noindent\textbf{Aggregation anomaly.}
FedIA improves FedAvg on S1/S3 (57.22$\to$60.72; 56.50$\to$59.82), but FGGP's S1 average barely changes (63.65$\to$63.83; Figure~\ref{fig:aggregation_comparison}). Beneath it, DE/ES/FR gain $+4.29/+5.20/+8.85$, EN loses 16.00, and PT/RU barely change (Appendix~\ref{app:full_results}). The average hides domain redistribution: prototype alignment reshapes useful support, after which a shared mask can over-filter a domain-specific remainder. FedHEAL's 63.65$\to$62.43 drop in the same block reinforces this boundary; strong representation alignment may require domain-adaptive selection beyond one shared mask.

\noindent\textbf{Mechanism and stability.}
FGGP's severe-skew deviation falls from 6.03 to 0.38; Figure~\ref{fig:fggp_convergence} attributes this to removing repeated late-round drops. IM supplies 2.58 of the full 3.50-point S1 gain, but is non-convergent on S3 unless CAM is added (Figure~\ref{fig:component_ablation}). Thus IM decides \textbf{where} to aggregate and CAM stabilizes \textbf{how much}. Appendix~\ref{app:additional_results} gives the per-domain ablation, $5\times5$ $\rho/\beta$ landscape, selected pairs, and bounded privacy verification from gradient inversion attacks.

\section{Conclusion}
\label{sec:conclusion}

This work dissects how graph-domain shift is expressed in client updates and transformed by server aggregation.
Salient update sharing erodes over training, cross-domain updates weakly reinforce the aggregate, and aggregation compatibility varies across graph-learning configurations and multiple domain factors.
These local patterns create component-dependent contributor sets, while ordinary averaging applies one client mixture to every component, producing aggregation-induced signal dilution.
FedIA addresses this mismatch through layer-wise parameter selection and momentum-smoothed client weighting without changing local graph learning or the standard update payload.
Across Twitch Gamers and WikiNet, every paired FedIA integration improves average accuracy and the gains extend across multiple domains; the same server rule transfers from propagation-deferred PMLP-GCN to message-passing GraphSAGE.
The resulting perspective places server aggregation alongside local graph learning as a measurable and actionable stage of multi-domain FGL.

\bibliography{aaai2027}

\clearpage
\begin{center}
\centering
{\Large\bfseries Supplementary Material of FedIA\par}
\end{center}
\appendix
\section{Experimental Protocol}\label{app:experiments}\label{app:exs}

\begin{table*}[!t]
    \centering
    \small
    \caption{Accuracy comparison on the \textbf{Twitch Gamers} cross-domain setting (Ratio 1:1:1:1:1:1, 12 Clients). Results are reported as mean$_{\pm\text{std}}$ over the last 20 rounds. Methods are grouped pairwise (baseline vs. enhanced) with alternating gray shading. \textbf{Annotations:} The column $\boldsymbol{\Delta}$ denotes the accuracy gap relative to generic FedAvg. Blue and red superscripts indicate gains and losses relative to the \textbf{original baseline} in each group.}
    \label{tab:twitch_cross-domain}
    
    \resizebox{\linewidth}{!}{\begin{tabular}{lcccccc|cc}
        \toprule
        \multirow{2}{*}{\textbf{Methods}} & \multicolumn{6}{c}{\textbf{Domains}} & \multicolumn{2}{c}{\textbf{Overall}} \\
        \cmidrule(lr){2-7} \cmidrule(lr){8-9}
         & ES & DE & FR & RU & EN & PT & AVG & $\Delta$\\
        \midrule
        
        FedAvg & \res{66.46}{0.02} & \res{38.38}{0.02} & \res{58.84}{0.03} & \res{68.74}{0.03} & \res{49.36}{0.02} & \res{61.57}{0.00} & \res{57.22}{0.01} & --\\
        \quad \textbf{+ FedIA} & \res{64.77}{0.05} & \res{55.90}{0.03} & \res{58.72}{0.03} & \res{65.41}{0.06} & \res{58.66}{0.04} & \res{60.85}{0.04} & \res{60.72}{0.02} & \imp{3.50}\gain{3.50}\\
        
        \rowcolor{grayrow} FedProx & \res{65.70}{0.00} & \res{38.20}{0.01} & \res{58.45}{0.00} & \res{68.37}{0.02} & \res{48.73}{0.01} & \res{61.31}{0.00} & \res{56.79}{0.00} & -0.43\\
        \rowcolor{grayrow} \quad \textbf{+ FedIA} & \res{70.81}{0.08} & \res{51.44}{0.14} & \res{60.07}{0.10} & \res{73.34}{0.10} & \res{53.28}{0.06} & \res{68.73}{0.09} & \res{62.95}{0.05} & \imp{5.73}\gain{6.16}\\

        MOON & \res{66.55}{0.02} & \res{38.37}{0.00} & \res{58.70}{0.02} & \res{68.95}{0.06} & \res{49.18}{0.02} & \res{61.46}{0.05} & \res{57.20}{0.01} & -0.02\\
        \quad \textbf{+ FedIA} & \res{70.48}{0.00} & \res{38.95}{0.00} & \res{62.23}{0.00} & \res{76.57}{0.00} & \res{46.05}{0.00} & \res{66.41}{0.00} & \res{60.11}{0.00} & \imp{2.89}\gain{2.91}\\

        \rowcolor{grayrow} FedDyn & \res{67.51}{0.34} & \res{38.86}{0.13} & \res{59.45}{0.21} & \res{70.83}{0.11} & \res{49.31}{0.30} & \res{63.65}{0.19} & \res{58.27}{0.14} & +1.05\\
        \rowcolor{grayrow} \quad \textbf{+ FedIA} & \res{69.81}{0.49} & \res{53.35}{0.63} & \res{61.28}{0.08} & \res{69.84}{0.11} & \res{56.12}{0.26} & \res{68.82}{0.40} & \res{63.20}{0.06} & \imp{5.98}\gain{4.93}\\
        
        FedProc & \res{66.46}{0.02} & \res{38.35}{0.01} & \res{58.74}{0.03} & \res{68.65}{0.03} & \res{49.29}{0.02} & \res{61.57}{0.00} & \res{57.18}{0.01} & -0.04\\
        \quad \textbf{+ FedIA} & \res{66.70}{0.20} & \res{54.68}{0.45} & \res{58.53}{0.08} & \res{63.26}{0.21} & \res{57.25}{0.07} & \res{66.08}{0.17} & \res{61.08}{0.03} & \imp{3.86}\gain{3.90}\\

        \rowcolor{grayrow} FedProto & \res{66.51}{0.00} & \res{38.35}{0.03} & \res{58.80}{0.01} & \res{68.67}{0.03} & \res{49.35}{0.03} & \res{61.57}{0.00} & \res{57.21}{0.00} & -0.01\\
        \rowcolor{grayrow} \quad \textbf{+ FedIA} & \res{68.19}{0.07} & \res{48.03}{0.13} & \res{59.46}{0.08} & \res{68.87}{0.08} & \res{55.28}{0.09} & \res{67.82}{0.15} & \res{61.27}{0.02} & \imp{4.05}\gain{4.06}\\

        FGSSL & \res{67.40}{0.05} & \res{38.84}{0.03} & \res{59.04}{0.02} & \res{69.01}{0.04} & \res{49.63}{0.03} & \res{62.17}{0.10} & \res{57.68}{0.02} & +0.48\\
        \quad \textbf{+ FedIA} & \res{68.27}{0.04} & \res{48.64}{0.02} & \res{60.29}{0.03} & \res{69.41}{0.05} & \res{54.03}{0.04} & \res{68.00}{0.09} & \res{61.44}{0.03} & \imp{4.22}\gain{3.76}\\

        \rowcolor{grayrow} FGGP & \res{67.98}{1.35} & \res{65.72}{0.37} & \res{56.85}{0.91} & \res{66.18}{1.35} & \res{56.12}{0.71} & \res{69.07}{0.41} & \res{63.65}{0.42} & +6.43\\
        \rowcolor{grayrow} \quad \textbf{+ FedIA} & \res{72.27}{0.50} & \res{49.72}{2.04} & \res{62.05}{0.39} & \res{75.03}{1.07} & \res{54.99}{5.20} & \res{68.92}{0.84} & \res{63.83}{1.19} & \imp{6.61}\gain{0.18}\\
        
        \bottomrule
    \end{tabular}}
\end{table*}

\begin{table*}[!t]
    \centering
    \small
    \caption{Accuracy comparison on the \textbf{Twitch Gamers} domain skew setting (Robust-verify Setting; Ratio 1:5:1:1:1:1, 20 Clients). Results are reported as mean$_{\pm\text{std}}$ over the last 20 rounds. Methods are grouped pairwise (baseline vs. enhanced) with alternating gray shading. \textbf{Annotations:} The column $\boldsymbol{\Delta}$ denotes the accuracy gap relative to generic FedAvg. Blue and red superscripts indicate gains and losses relative to the \textbf{original baseline} in each group.}
    \label{tab:comptwitch}
    
    \resizebox{\linewidth}{!}{\begin{tabular}{lcccccc|cc}
        \toprule
        \multirow{2}{*}{\textbf{Methods}} & \multicolumn{6}{c}{\textbf{Domains}} & \multicolumn{2}{c}{\textbf{Overall}} \\
        \cmidrule(lr){2-7} \cmidrule(lr){8-9}
         & ES & DE & FR & RU & EN & PT & AVG & $\Delta$\\
        \midrule
        
        FedAvg & \res{67.14}{0.05} & \res{38.70}{0.01} & \res{58.76}{0.00} & \res{65.95}{0.03} & \res{49.05}{0.03} & \res{59.48}{0.00} & \res{56.51}{0.01} & --\\
        \quad \textbf{+ FedIA} & \res{67.80}{0.09} & \res{55.51}{0.05} & \res{59.61}{0.02} & \res{64.31}{0.05} & \res{57.45}{0.02} & \res{59.61}{0.02} & \res{61.31}{0.02} & \imp{4.80}\gain{4.80}\\
        
        \rowcolor{grayrow} FedProx & \res{66.61}{0.00} & \res{38.39}{0.00} & \res{58.64}{0.00} & \res{66.07}{0.02} & \res{59.08}{0.00} & \res{59.08}{0.00} & \res{56.13}{0.00} & -0.38\\
        \rowcolor{grayrow} \quad \textbf{+ FedIA} & \res{70.44}{0.18} & \res{54.41}{0.08} & \res{59.06}{0.10} & \res{68.13}{0.08} & \res{57.71}{0.06} & \res{59.06}{0.10} & \res{62.96}{0.03} & \imp{6.45}\gain{6.83}\\

        MOON & \res{67.03}{0.03} & \res{38.77}{0.01} & \res{58.97}{0.02} & \res{66.07}{0.02} & \res{48.72}{0.03} & \res{59.35}{0.03} & \res{56.49}{0.01} & -0.02\\
        \quad \textbf{+ FedIA} & \res{71.11}{0.07} & \res{41.15}{0.03} & \res{63.33}{0.03} & \res{70.05}{0.04} & \res{50.16}{0.06} & \res{63.33}{0.03} & \res{60.04}{0.02} & \imp{3.53}\gain{3.55}\\

        \rowcolor{grayrow} FedDyn & \res{50.00}{21.94} & \res{50.00}{11.05} & \res{50.00}{14.02} & \res{50.06}{24.00} & \res{50.00}{4.47} & \res{50.00}{14.31} & \res{50.01}{9.79} & -6.50\\
        \rowcolor{grayrow} \quad \textbf{+ FedIA} & \res{68.16}{0.57} & \res{55.14}{0.44} & \res{58.13}{0.85} & \res{66.74}{0.48} & \res{56.25}{0.90} & \res{58.13}{0.85} & \res{61.42}{0.31} & \imp{4.91}\gain{11.41}\\
        
        FedProc & \res{67.04}{0.03} & \res{38.72}{0.02} & \res{58.77}{0.03} & \res{65.94}{0.03} & \res{48.88}{0.02} & \res{59.41}{0.07} & \res{56.46}{0.01} & -0.05\\
        \quad \textbf{+ FedIA} & \res{64.36}{0.04} & \res{60.05}{0.02} & \res{55.28}{0.03} & \res{62.09}{0.08} & \res{57.64}{0.02} & \res{67.99}{0.04} & \res{61.23}{0.01} & \imp{4.72}\gain{4.77}\\

        \rowcolor{grayrow} FedProto & \res{67.15}{0.02} & \res{38.66}{0.02} & \res{58.71}{0.02} & \res{65.96}{0.00} & \res{48.98}{0.02} & \res{59.46}{0.04} & \res{56.48}{0.01} & -0.03\\
        \rowcolor{grayrow} \quad \textbf{+ FedIA} & \res{74.06}{0.00} & \res{38.95}{0.00} & \res{64.02}{0.00} & \res{71.94}{0.00} & \res{45.53}{0.00} & \res{64.02}{0.00} & \res{59.80}{0.00} & \imp{3.29}\gain{3.32}\\

        FGSSL & \res{67.71}{0.04} & \res{39.20}{0.03} & \res{58.91}{0.04} & \res{65.73}{0.09} & \res{49.50}{0.04} & \res{59.83}{0.07} & \res{56.81}{0.02} & +0.30\\
        \quad \textbf{+ FedIA} & \res{64.36}{0.21} & \res{56.01}{0.13} & \res{56.93}{0.04} & \res{63.91}{0.11} & \res{56.91}{0.21} & \res{56.93}{0.04} & \res{59.82}{0.05} & \imp{3.31}\gain{3.01}\\

        \rowcolor{grayrow} FGGP & \res{65.99}{2.35} & \res{66.28}{0.76} & \res{55.41}{1.95} & \res{63.97}{3.10} & \res{55.70}{1.58} & \res{67.89}{2.34} & \res{62.54}{1.24} & +6.03\\
        \rowcolor{grayrow} \quad \textbf{+ FedIA} & \res{68.86}{1.28} & \res{69.69}{2.52} & \res{51.22}{1.77} & \res{62.78}{2.04} & \res{60.01}{1.58} & \res{72.50}{0.58} & \res{64.17}{0.30} & \imp{7.66}\gain{1.63}\\
        
        \bottomrule
    \end{tabular}}
\end{table*}

\begin{table*}[t]
\centering
\scriptsize
\caption{Full domain-level ablation on Twitch Gamers. Results are mean$\pm$std accuracy (\%). IM denotes Importance Masking and CAM denotes contribution-aware momentum.}
\label{tab:app_ablation}
\resizebox{\linewidth}{!}{\begin{tabular}{llccccccc}
\toprule
\textbf{Setting} & \textbf{Method} & ES & DE & FR & RU & EN & PT & AVG\\
\midrule
S1 & FedAvg & \res{66.46}{0.02} & \res{38.38}{0.02} & \res{58.84}{0.03} & \res{68.74}{0.03} & \res{49.36}{0.02} & \res{61.57}{0.00} & \res{57.22}{0.01}\\
S1 & FedAvg + IM & \res{71.29}{0.00}\gain{4.83} & \res{38.95}{0.02}\gain{0.57} & \res{64.02}{0.00}\gain{5.18} & \res{76.23}{0.05}\gain{7.49} & \res{45.54}{0.02}\loss{3.82} & \res{62.76}{0.09}\gain{1.19} & \res{59.80}{0.01}\gain{2.58}\\
S1 & FedAvg + IM + CAM & \res{64.77}{0.05}\loss{1.69} & \res{55.90}{0.03}\gain{17.52} & \res{58.72}{0.03}\loss{0.12} & \res{65.41}{0.06}\loss{3.33} & \res{58.66}{0.04}\gain{9.30} & \res{60.85}{0.04}\loss{0.72} & \res{60.72}{0.02}\gain{3.50}\\
\midrule
S3 & FedAvg & \res{66.98}{0.06} & \res{38.76}{0.01} & \res{59.38}{0.05} & \res{65.06}{0.02} & \res{48.87}{0.03} & \res{59.97}{0.10} & \res{56.50}{0.04}\\
S3 & FedAvg + IM & NC & NC & NC & NC & NC & NC & NC\\
S3 & FedAvg + IM + CAM & \res{71.29}{0.00}\gain{4.31} & \res{38.95}{0.00}\gain{0.19} & \res{62.75}{0.00}\gain{3.37} & \res{76.40}{0.00}\gain{11.34} & \res{45.53}{0.00}\loss{3.34} & \res{64.02}{0.00}\gain{4.05} & \res{59.82}{0.00}\gain{3.32}\\
\bottomrule
\end{tabular}}
\end{table*}

\begin{figure}[t]
\centering
\includegraphics[width=\linewidth]{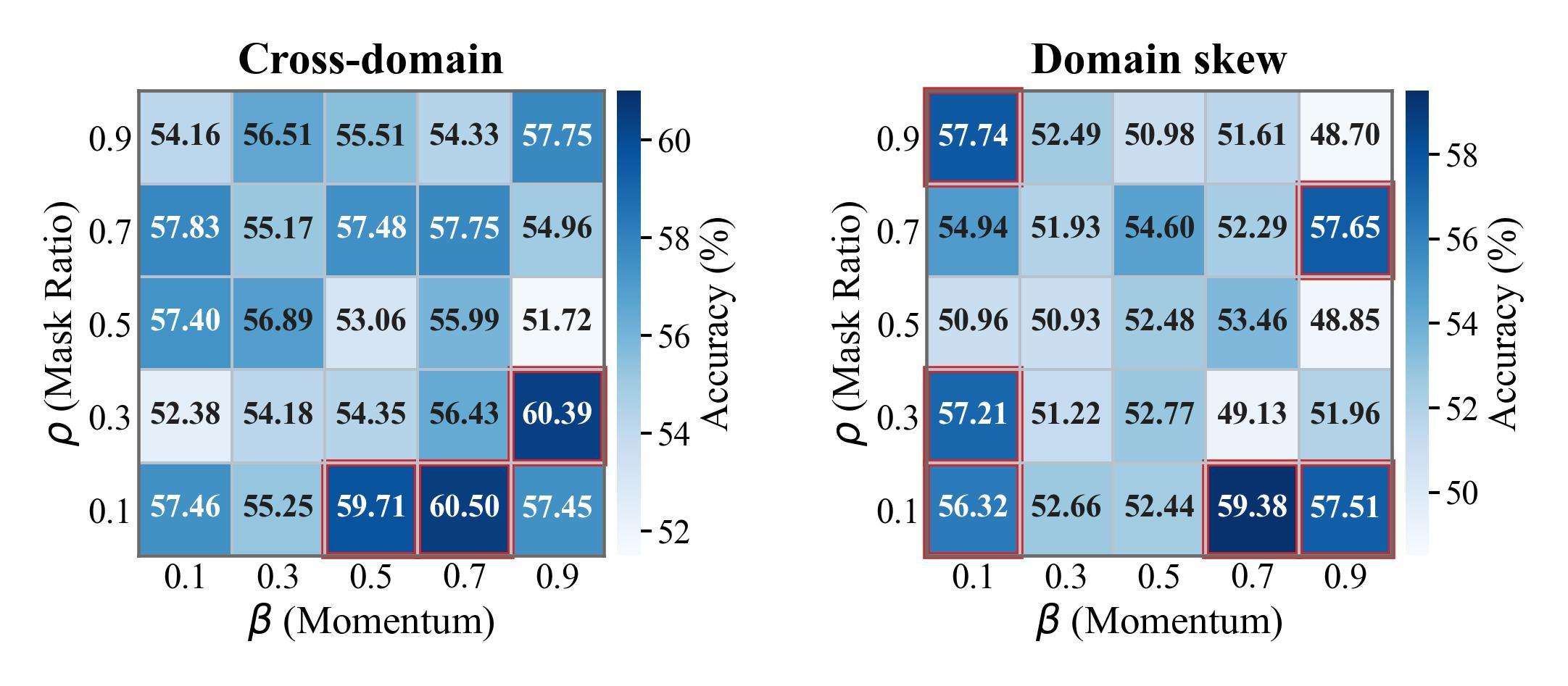}
\caption{Hyperparameter sensitivity of FedIA across the $(\rho,\beta)$ grid. The left and right panels show the cross-domain S1 and severe-skew S3 settings; red boxes indicate regions that exceed the corresponding average baseline.}
\label{fig:app_hyperparameters}
\end{figure}

\begin{table}[t]
\centering
\scriptsize
\caption{Selected $(\rho,\beta)$ pairs for FedIA across Twitch settings.}
\label{tab:app_hyperparameters}
\resizebox{\linewidth}{!}{\begin{tabular}{lcccccc}
\toprule
\textbf{Base method} & \multicolumn{2}{c}{S1} & \multicolumn{2}{c}{S2} & \multicolumn{2}{c}{S3}\\
 & $\rho$ & $\beta$ & $\rho$ & $\beta$ & $\rho$ & $\beta$\\
\midrule
FedAvg & 0.1 & 0.1 & 0.1 & 0.3 & 0.3 & 0.3\\
FedProx & 0.9 & 0.3 & 0.7 & 0.5 & 0.5 & 0.7\\
MOON & 0.3 & 0.1 & 0.1 & 0.9 & 0.1 & 0.9\\
FedDyn & 0.3 & 0.9 & 0.5 & 0.3 & 0.1 & 0.1\\
FedProto & 0.1 & 0.3 & 0.3 & 0.1 & 0.1 & 0.9\\
FGSSL & 0.1 & 0.1 & 0.1 & 0.1 & 0.1 & 0.3\\
FGGP & 0.3 & 0.5 & 0.5 & 0.3 & 0.3 & 0.9\\
\bottomrule
\end{tabular}}
\end{table}

In this supplementary material, we provide reproducibility details and supplementary accuracy tables that are not repeated in the main text.

We begin by presenting the \textbf{Detailed Setup}, which outlines our computing environment, the rationale for dataset selection, data partitioning strategies for creating cross-domain and domain-skew scenarios, and comprehensive descriptions of all backbone models and baseline methods.

Appendix B reports additional Twitch Gamers results, full main-domain tables, and aggregation compatibility tests that complement the compressed main performance section.

Appendix C reports the full ablation, hyper-parameter robustness, and bounded privacy stress-test details needed to interpret the main-paper summary.

\subsection{Detailed Setup}

We conducted experiments using NVIDIA GeForce RTX 4080 hardware, CUDA Driver 12.6, Driver Version 561.17, and Python 3.11. Each client performs five local iterations with SGD~\citep{sgd}, momentum $0.9$, weight decay $10^{-5}$, and learning rate $\eta=0.05$ for 200 communication rounds. Each trajectory is summarized over its final 20 rounds. Further details follow.

\subsubsection{Datasets}

To achieve a fair and effective comparison in our study, we selected the dataset based on the following rationale: (1) To enhance the experimental reliability, we opted for datasets that are publicly available, sourced from the same interface, yet encompass different domains. (2) To minimize the effects of model heterogeneity, we chose datasets with consistent feature dimensions and label categories. (3) The dataset inherently exhibits cross-domain characteristics, enabling the development of domain skew experimental scenarios through random partitioning. Taking these factors into account, we chose the Twitch Gamers Dataset (e.g. \textbf{Twitch}) \citep{twitch} and Wikipedia Network Dataset (e.g. \textbf{WikiNet}) \citep{twitch} from the homogeneity collection of PyG library \citep{pyg}.

The \textbf{Twitch} Gamer networks dataset encompasses six domains: DE, EN, ES, FR, PT, and RU. Each domain features 128-dimensional node attributes, with all nodes representing binary classifications. Specifically, the DE domain comprises 9,498 nodes and 315,774 edges; the EN domain includes 7,126 nodes and 77,774 edges; the ES domain consists of 4,648 nodes and 123,412 edges; the FR domain contains 6,551 nodes and 231,883 edges; the PT domain features 1,912 nodes and 64,510 edges; and the RU domain has 4,385 nodes and 78,993 edges. Given that the DE domain has the highest number of nodes and edges, our design for addressing domain skew is primarily focused on this domain.

The \textbf{Wiki}pedia \textbf{Net}works dataset collection includes three networks: chameleon, crocodile, and squirrel. In these networks, nodes represent Wikipedia articles, and edges indicate mutual links between them. Each network features multi-dimensional node attributes derived from bag-of-words representations of the articles, and all nodes are classified into one of five categories. Specifically, the chameleon network comprises 2,277 nodes and 36,101 edges, with 2,089-dimensional features. The crocodile network consists of 11,631 nodes and 170,918 edges, with 2,089-dimensional features. The squirrel network contains 5,201 nodes and 217,073 edges, also with 2,089-dimensional features. Given that the squirrel network has the highest number of edges, our design for addressing network density would be primarily focused on this network.

For the cross-domain setting, Twitch S1 uses 12 clients and WikiNet uses 6 clients. For domain skew, the Twitch DE graph is randomly partitioned to create fivefold and tenfold client allocations relative to each other domain, yielding S2 with 20 clients and S3 with 30 clients. Twitch nodes are split into train/validation/test sets at $60\%/10\%/30\%$, while WikiNet uses $60\%/20\%/20\%$. All experiments use full participation.

\subsubsection{Baselines \& Backbone}

We employed \textbf{PMLP-GCN}, utilize the 2 layers feature extractor and set the hidden layer size of 128, as the backbone for our major experiments \citep{pmlp}; and employed \textbf{GraphSAGE}‑mean \citep{graphsage}, utilizing a 2‑layer neighborhood aggregation (K = 2, fan‑outs of 25 and 10) and set the hidden embedding dimension to 128 as another backbone for our FedSage Comparison experiments. The baseline methods we utilized include:
\textbf{FedAvg} \citep{fedavg} iteratively averages client model parameters to form a global model, serving as the classical baseline for communication-efficient federated optimization.
\textbf{FedProx} \citep{fedprox} introduces a proximal term into each client’s objective to curb local drift, yielding stabler convergence under pronounced systems and statistical heterogeneity. 
\textbf{Moon} \citep{moon} leverages model-level contrastive learning between local and global representations to align feature spaces and counteract non-IID data effects. 
\textbf{FedDyn} \citep{feddyn} adds a dynamic regularizer—effectively a control variate—that steers local optima toward the global objective, accelerating convergence without extra communication. 
\textbf{FedOPT} \citep{fedopt} generalizes FedAvg by applying adaptive server-side optimizers (e.g., Adam, Yogi) to aggregated updates, improving learning speed and robustness on diverse clients. 
\textbf{FedProc} \citep{fedproc} employs a prototypical contrastive loss that distills global class prototypes to guide local training, narrowing performance gaps on highly non-IID data. 
\textbf{FedProto} \citep{fedproto} replaces gradient exchange with communication of class prototypes, which the server aggregates and redistributes to regularize heterogeneous client models. 
\textbf{FGSSL} \citep{fgssl} tackles federated graph learning by jointly aligning node semantics and graph structures through self-supervised global–local contrast, enabling label-efficient training.
\textbf{FGGP} \citep{fggp} learns domain-generalizable prototypes that decouple global and local representations, boosting cross-domain transferability in federated graph settings. 
It is important to note that we did not include \textbf{FedSSP} as a baseline in extensive experiments because FedSSP \citep{fedssp} follows the GSP methodology. Considering FedSage is an important baseline, we compare the performance with FedSage on GraphSage. This decision was made to maintain fairness in our comparisons.

\subsubsection{Hyper-parameters}

The hyperparameters of the FedIA framework include the mask ratio $\rho$ and the momentum factor $\beta$. We defined the search space for both hyperparameters as $\{0.1, 0.3, 0.5, 0.7, 0.9\}$. The selected values and sensitivity analysis are reported in the main paper; this supplement records the search space for reproducibility.

\section{Complete Performance Results}\label{app:full_results}

\subsection{Additional Twitch Settings}

In this section, we report supplementary Twitch Gamers results that contextualize the main performance comparison.

\subsubsection{Twitch Performance Analysis}
\label{sec:appendix_twitch_analysis}

To supplement the main results, we report Twitch Gamers performance under two regimes: the balanced \textit{Cross-domain} setting (Table~\ref{tab:twitch_cross-domain}) and the moderate \textit{Domain-skew} setting (Table~\ref{tab:comptwitch}).

\textbf{Performance under Balanced Cross-domain Shifts (Table~\ref{tab:twitch_cross-domain}).} 
In this scenario (12 clients, Ratio 1:1:1:1:1:1), data are evenly distributed across six regions. As shown in Table~\ref{tab:twitch_cross-domain}, FedIA improves the average accuracy for all tested baseline groups, with gains ranging from \textbf{+0.18\%} to \textbf{+6.16\%}.
\textbf{(1) General FL optimizers:} FedProx and FedDyn gain \textbf{+6.16\%} and \textbf{+4.93\%} over their original baselines, respectively, indicating that aggregation-side calibration can complement regularization-based optimizers under graph-domain shifts.
\textbf{(2) Specialized FGL methods:} FedIA also remains compatible with graph-specific methods. For example, FedProc improves by \textbf{+3.90\%}, while FGGP+FedIA reaches the highest overall accuracy of \textbf{63.83\%}, a modest \textbf{+0.18\%} over FGGP and \textbf{+6.61\%} relative to FedAvg.

\textbf{Resilience under Moderate Domain Skew (Table~\ref{tab:comptwitch}).} 
Table~\ref{tab:comptwitch} presents a more challenging moderate domain-skew setting (20 clients, Ratio 1:5:1:1:1:1), where the DE domain dominates the network. FedIA yields positive gains across the tested baseline groups:
\textbf{(1) Recovery of unstable strategies:} A notable result is observed for FedDyn. While the vanilla FedDyn baseline drops to 50.01\% accuracy under this skew, FedDyn+FedIA reaches 61.42\%, corresponding to \textbf{+11.41\%} over FedDyn and \textbf{+4.91\%} relative to FedAvg.
\textbf{(2) Improvements across baselines:} FedIA also improves MOON, FedProc, and FedProto by \textbf{+3.55\%}, \textbf{+4.77\%}, and \textbf{+3.32\%} over their original baselines, respectively.
\textbf{(3) Strongest overall accuracy:} FGGP+FedIA reaches the highest accuracy of \textbf{64.17\%}, a \textbf{+1.63\%} gain over FGGP and \textbf{+7.66\%} relative to FedAvg.

\textbf{Summary.} 
The results across Tables~\ref{tab:twitch_cross-domain} and~\ref{tab:comptwitch} support a bounded conclusion: FedIA's two-stage server-side calibration improves aggregation for the evaluated Twitch cross-domain and domain-skew settings, with larger gains for unstable or low-performing baselines and more modest gains for already strong graph-specific methods.

\subsubsection{Additional Accuracy Tables}
\label{sec:appendix_tables}

The following tables provide supplementary Twitch accuracy results not shown as main performance tables.
Table~\ref{tab:twitch_cross-domain} reports the cross-domain setting (12 clients, balanced split).
Table~\ref{tab:comptwitch} reports the moderate domain-skew setting (20 clients, 5-fold split).
The 30-client severe domain-skew setting and the compact diagnostic figures are reported in the main text; Appendix C retains the full ablation and hyper-parameter robustness details.

\section{Component and Robustness Analysis}\label{app:additional_results}

This appendix records the detailed evidence behind the diagnostics in Figures~\ref{fig:fggp_convergence}, \ref{fig:aggregation_comparison}, and \ref{fig:component_ablation}: the full domain-level ablation, the $\rho/\beta$ sensitivity analysis, and the bounded privacy stress test.

\subsection{Full Ablation}

The full domain-level ablation separates Importance Masking (IM), contribution-aware momentum (CAM), and the complete FedIA combination. The S3 IM-only row is marked as non-convergent, as in Figure~\ref{fig:component_ablation}.

\subsection{Hyperparameter Sensitivity}

Figure~\ref{fig:app_hyperparameters} reports the full $\rho$/$\beta$ heatmaps for S1 and S3. These heatmaps are descriptive robustness evidence, not a separate main efficacy claim. The selected pairs in Table~\ref{tab:app_hyperparameters} document the search outcome used for the reported settings.

\subsection{Privacy Stress Test}

Gradient inversion is a known risk in gradient-sharing systems, including DLG~\citep{dlg}, and has recently been strengthened for graph data by GRAIN~\citep{drencheva2025grain}. We conduct a focused stress test to check whether FedIA's server-side calibration introduces additional inversion risk compared with the same pipeline without FedIA. Under the evaluated GRAIN setting across 62 runs, the measured privacy score shifts from 0.519 to 0.498, a 4.0\% reduction. This bounded empirical check suggests that FedIA does not increase the measured reconstruction score under the tested configuration; it is not a formal privacy guarantee, and a complete threat analysis remains beyond the scope of this work.

\end{document}